\documentclass[final,numbered]{ifacconf}

\usepackage{caption}
\usepackage[strings]{underscore}
\usepackage{graphicx}
\usepackage{natbib}
\usepackage{amsmath}
\usepackage{amsfonts}
\usepackage{booktabs}
\usepackage[compatibility=false]{caption}
\usepackage{float}
\usepackage{adjustbox}

\begin{document}

\begin{frontmatter}

\title{Explainability-Driven Feature Engineering for Mid-Term Electricity Load Forecasting in ERCOT's SCENT Region}

\author{Abhiram Bhupatiraju}
\address{Institute for Computing in Research}

\author{Sung Bum Ahn}
\address{Electric Reliability Council of Texas, Taylor TX}

\begin{abstract}
Accurate load forecasting is essential to the operation of modern electric power systems. Given the sensitivity of electricity demand to weather variability and temporal dynamics, capturing non-linear patterns is essential for long-term planning. This paper presents a comparative analysis of machine learning models, Linear Regression, XGBoost, LightGBM, and Long Short-Term Memory (LSTM), for forecasting system-wide electricity load up to one year in advance. Midterm forecasting has shown to be crucial for maintenance scheduling, resource allocation, financial forecasting, and market participation. The paper places a focus on the use of a method called  "Shapley Additive Explanations" (SHAP) to improve model explainability. SHAP enables the quantification of feature contributions, guiding informed feature engineering and improving both model transparency and forecasting accuracy.\end{abstract}

\end{frontmatter}

\section*{Keywords and Abbreviations}

\begin{center}
\footnotesize
\begin{tabular}{ll}
\textbf{ML} & Machine Learning \\
\textbf{AI} & Artificial Intelligence \\
\textbf{ELF} & Electric Load Forecasting \\
\textbf{XAI} & Explainable Artificial Intelligence \\
\textbf{SHAP} & SHapley Additive Explanations \\
\textbf{LSTM} & Long Short-Term Memory \\
\textbf{CNN} & Convolutional Neural Network \\
\textbf{XGB} & eXtreme Gradient Boosting \\
\textbf{LR} & Linear Regression \\
\textbf{LightGBM} & Light Gradient Boosting Machine \\
\end{tabular}
\hspace{2cm}
\begin{tabular}{ll}
\textbf{PLF} &  \hspace{1.2em}Probabilistic Load Forecasting \\
\textbf{RNN} & \hspace{1.2em}Recurrent Neural Network \\
\textbf{FCM} & \hspace{1.2em}Fuzzy Cognitive Maps \\
\textbf{AR} & \hspace{1.2em}Autoregressive Model \\
\textbf{SVM} & \hspace{1.2em}Support Vector Machine \\
\textbf{MAE} & \hspace{1.2em}Mean Absolute Error \\
\textbf{MAPE} & \hspace{1.2em}Mean Absolute Percentage Error \\
\textbf{RMSE} & \hspace{1.2em}Root Mean Square Error \\
\textbf{ERCOT} & \hspace{1.2em}Electric Reliability Council of Texas \\
\textbf{SCENT} & \hspace{1.2em}South Central Texas Region \\
\end{tabular}
\end{center}
\section{Introduction}

Load forecasting is a critical component for modern grid operation and planning. As global energy systems transition toward increased integration of renewable energy sources, face increasing demand, and even the proliferation of large data centers, grid operators must anticipate fluctuating demand with increased precision. Inaccurate forecasts can lead to inefficient scheduling, higher operational costs and even grid instability during peak load conditions. With snow storms and heatwaves becoming even more unpredictable as time goes on, the need for robust forecasting models increases in the context of smart grids, where system flexibility and ability to grasp non-linear patterns grows paramount.

Mid-term load forecasting (MTLF), which generally refers to forecast horizons ranging from one week to one year, plays a unique role within the broader forecast spectrum. Although short-term forecasting is essential for real-time grid operations and long-term forecasting supports infrastructure development, MTLF is indispensable for maintenance scheduling, financial planning, energy generation, and regulatory compliance \cite{yin2023mid}. Unlike Short-Term Forecasting it remains more volatile  due to seasonal, behavioral, and climatic uncertainty. Traditional statistical models, such as ARIMA and exponential smoothing, often struggle with the non-linear and multi-scale patterns inherent to MTLF, creating space for the application of machine learning approaches. \cite{ghiassi2006medium, yin2023mid}.

Recent advances in machine learning (ML) and deep learning (DL) have significantly improved prediction in MTLF applications. These models are well suited to capture nonlinear relationships, temporal dependencies, and high-dimensional feature interactions that traditional approaches fail to model effectively \cite{yin2023mid, waheed2024empowering}. Techniques such as linear regression and ARIMA cannot be obtained. On the other hand, gradient-boosted trees (e.g. XGBoost and LightGBM) and recurrent neural networks (e.g. LSTM) have demonstrated superior predictive accuracy across various time horizons and geographies \cite{waheed2024empowering, ghiassi2006medium}. Despite their empirical success, the practical deployment of these models is often hindered by their "black-box" nature, which limits operator trust, reduces transparency, and impairs the ability to effectively feature engineer. Adding models with greater complexity was found to be inversely proportional to ideas such as explainability \cite{baur2024explainability}.

To address these concerns, the field has seen growing interest in explainable artificial intelligence (XAI) techniques that aim to make ML/DL models more transparent. Among them, SHAP (SHapley Additive exPlanations) has emerged as a leading framework for interpreting model outputs, offering global and local explanations of feature contributions \cite{baur2024explainability, neubauer2025explainableb}. Recent research highlights the importance of integrating interpretability into load forecast pipelines to support human decision making and improve model diagnostics \cite{neubauer2025explainableb, baur2024explainability}. However, many existing studies stop at model interpretation and do not take advantage of explainability to improve forecast accuracy, particularly during peak load periods, where precise predictions are most crucial.

Explainability is one of the most critical properties for machine learning models used in high-stakes domains, such as power systems. As defined by  \cite{baur2024explainability}, explainability pertains to the extent to which the internal mechanics of a model can be understood and communicated. Increasing explainability is necessary to improve model performance and understanding feature importance. Despite the growing use of ML in load forecasting, many studies treat these models as black boxes, limiting their adoption in the real world \cite{baur2024explainability}.

Our work builds on this foundation by adopting SHAP, a model-agnostic explainability method rooted in cooperative game theory. Originally developed to fairly attribute payouts to players in a game, SHAP has been adapted to machine learning to quantify the contribution of each feature to a prediction using Shapley values. In the context of electricity load forecasting, SHAP enables not only post hoc explainability but also deeper diagnostic insights into model behavior. It is able to do various calculations to provide a numerical value that each variable or feature has contributed to the prediction. In this study, we leverage SHAP to identify the most influential characteristics during extreme peak demand events, traditionally the most difficult forecast scenarios. By using these insights we are able to understand which features work positively to predict peak demands and which hurt our predictions. We are also able to engineer new features and refine the modeling pipeline. Additionally, our study aims to show that SHAP can be used not just for transparency, but also as a tool to improve mid-term forecast accuracy where it matters most.

Midterm load forecasting has evolved through complementary advances in modeling techniques, explainability tools, and feature engineering. Our work builds upon these efforts by using SHAP explainability not only for interpretation but also as a tool for targeted feature engineering—particularly for improving peak load forecasts, which remain underexplored in existing literature.

\subsection{Foundational MTLF Models: From Deep Learning to Hybrid Architectures}

Oreshkin et al. introduced the \textit{N-BEATS} architecture for time series forecasting, replacing classical seasonality modeling with deep residual blocks that learn patterns directly from data \cite{Oreshkin2020NBEATSNN}. Its successor, \textit{N-BEATS*} \cite{Kasprzyk2024EnhancedNF}, improved accuracy through a novel loss function and post-processing layer. While both enhance midterm accuracy over large regions, they remain uni variate and cannot incorporate exogenous drivers such as temperature or holiday effects—an essential limitation for electricity demand modeling.

Zhang et al. proposed a hybrid framework combining seasonal decomposition with LSTM and XGBoost to capture linear trends and nonlinear residuals \cite{Kasprzyk2024EnhancedNF}. Similarly, Pełka and Dudek introduced pattern-based LSTMs that encode structural seasonality into symbolic forms to stabilize long-horizon forecasts \cite{Peka2020PatternbasedLS}. These models marked a shift toward hybrid architectures but lacked explainability or diagnostic tools to guide feature design.

Zimmermann and Ziel proposed a Generalized Additive Model (GAM) incorporating holiday-aware splines and autoregressive terms for interpretable midterm forecasting across Europe \cite{Zimmermann2024EfficientMF}. While their method improves interpretability and robustness, it primarily addresses national-level data and overlooks region-specific peak volatility and sub-hourly demand spikes.

\subsection{The Rise of Explainability in Load Forecasting}

Explainability tools have grown in importance alongside black-box models such as XGBoost and LSTM, especially in critical infrastructure contexts. Wu et al. used SHAP to interpret a multi-energy LSTM model, evaluating the influence of features across locations \cite{Wu2022AnEF}. However, SHAP was only applied post hoc and did not guide model training or feature construction.

Li and Wang developed GS-XGBoost for enterprise load forecasting and used SHAP to rank feature importance \cite{Li2022PowerLF}. While their approach improved model transparency, it did not establish an iterative feedback loop between SHAP insights and feature refinement. Furthermore, neither of these studies addressed peak load prediction—an operationally vital yet underexamined failure point.

Baur et al. systematically reviewed the challenges in explainable load forecasting, emphasizing SHAP’s potential for model diagnostics \cite{baur2024explainability}. They explicitly call for research that goes beyond interpretation and leverages explainability for performance improvements during critical events, such as heatwaves or winter storms.

Neubauer et al. partially answered this call by using SHAP and a temporal attention mechanism for heating load forecasting \cite{neubauer2025explainableb}. However, their improvement from explainability was not quantified, and their focus remained on heating systems, not electricity grids.

\subsection{Building Connections: What Is Missing and How This Work Extends It}

While prior work has advanced modeling (e.g., N-BEATS, hybrid LSTMs), transparency (e.g., GS-XGBoost, SHAP-enhanced LSTMs), and interpretability frameworks (e.g., Baur et al., Neubauer et al.), no prior study integrates these components into a unified pipeline. Specifically:

\begin{itemize}
    \item Deep learning models, though powerful, are often opaque and ignore exogenous variables (e.g., \cite{Oreshkin2020NBEATSNN, Kasprzyk2024EnhancedNF}).
    \item SHAP is rarely used to improve models iteratively; it is typically applied after training without informing feature engineering \cite{Li2022PowerLF, Wu2022AnEF}.
    \item Peak demand periods—operationally the most crucial—are seldom targeted explicitly.
    \item No previous work quantifies improvements in MAPE during the top 5\% of load hours from SHAP-inspired feature refinement.
\end{itemize}

\subsection{Contribution of This Paper}

Our work addresses these gaps by tightly integrating SHAP into the forecasting pipeline—not only to interpret model behavior but to iteratively improve it. Our contributions include:

\begin{itemize}
    \item Using SHAP to identify feature deficiencies during peak under prediction, leading to the creation of features such as \texttt{load\_spike\_vs\_mean} and \texttt{CDD\_x\_hour}.
    \item Evaluating models not only globally but also locally during peak windows, introducing a novel metric space often overlooked in past research.
    \item Demonstrating how SHAP importance rankings evolve after feature engineering and quantifying the resulting improvements—achieving a 3–6x reduction in MAPE during peak periods.
\end{itemize}

To the best of our knowledge, this is the first study to operationalize SHAP as a feedback loop for feature engineering in midterm electricity load forecasting. Our approach yields measurable improvements in both accuracy and transparency during peak demand events, bridging previously disconnected research directions in model interpretability and performance optimization.

\section{Machine Learning Models}
\subsection{A. Linear Regression}
Linear Regression is a fundamental statistical method used to model the relationship between a dependent variable and one or more independent variables. In its basic form, linear regression assumes that the relationship between the input variables $X$ and the target variable $y$ is linear:\cite{hastie2009elements}:

\begin{equation}
y = \beta_0 + \beta_1 x_1 + \beta_2 x_2 + \dots + \beta_n x_n + \varepsilon
\label{eq:linear_regression}
\cite{hastie2009elements}:
\end{equation}

Here, $\beta_0$ is the intercept, $\beta_1$ through $\beta_n$ are the coefficients learned from the data, and $\varepsilon$ is the error term. The model is trained using Ordinary Least Squares (OLS), which minimizes the sum of squared differences between the actual and predicted values. 

\begin{figure}[h]
\centering
\includegraphics[width=0.45\textwidth]{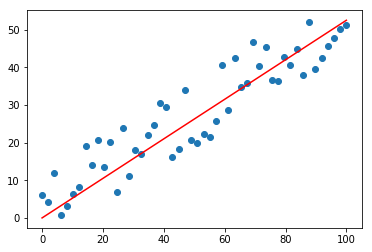}
\caption{A fitted linear regression model.}
\label{fig:linear}

\end{figure}

While linear regression is simple and interpretable, it may not capture complex, non-linear dependencies inherent in electricity load forecasting. Figure~\ref{fig:linear} illustrates a fitted linear regression model and its residuals.

\subsection{B. XGBoost (Extreme Gradient Boosting)}
\begin{figure}[h]
  \centering
  \includegraphics[width=0.45\textwidth]{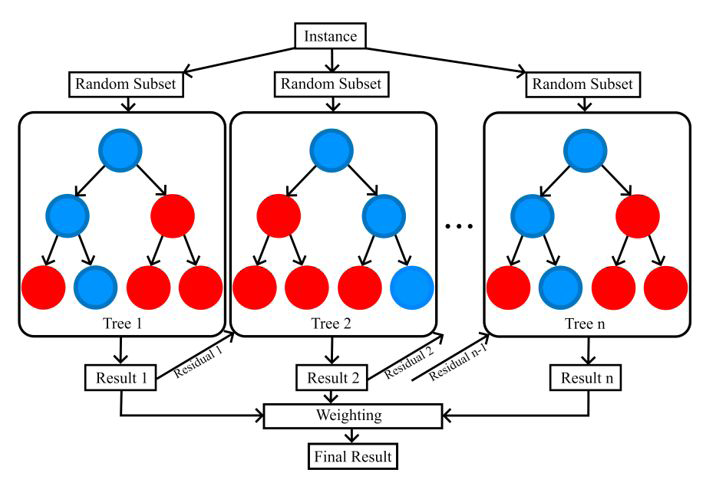}
  \caption{XGBoost Architecture Overview \cite{Oztornaci2025}}
  \label{fig:xgboost}
\end{figure}

XGBoost is a tree-based ensemble learning method that builds models in a sequential manner. Each new model attempts to correct errors made by previous ones using gradient descent optimization. It minimizes the regularized objective function\cite{chen2016xgboost}:
\begin{equation}
\mathcal{L}(\phi) = \sum\_{i} l(y\_i, \hat{y}\_i) + \sum\_k \Omega(f\_k)
\end{equation}
\cite{chen2016xgboost}:
where $l$ is a loss function (for example, mean square error), $\Omega(f_k)$ is a regularization term that penalizes complexity, and $\phi$ represents the model parameters. XGBoost incorporates shrinkage, column sampling, and tree pruning for better generalization. It handles missing values automatically and supports parallel computation. Figure~\ref{fig:xgboost} provides a visual overview of the XGBoost architecture.

\subsection{C. LightGBM (Light Gradient Boosting Machine)}
\begin{figure}[h]
  \centering
  \includegraphics[width=0.45\textwidth]{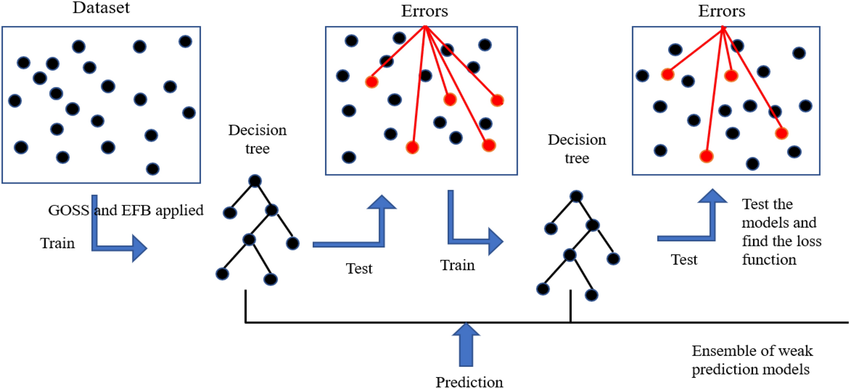}
  \caption{LightGBM Architecture Overview \cite{Deborah2024}}
  \label{fig:lightgmb}
\end{figure}

LightGBM is a gradient boost framework that uses histogram-based algorithms for faster training and lower memory usage compared to traditional tree-based models. It employs leaf-wise tree growth rather than level-wise, splitting the leaf with the maximum loss reduction. The objective is similar to XGBoost\cite{ke2017lightgbm}:
\begin{equation}
\mathcal{L} = \sum\_{i} l(y\_i, \hat{y}\_i) + \lambda \sum\_k \Omega(f\_k)
\end{equation}
LightGBM is particularly effective on large datasets and supports categorical feature handling natively, making it efficient and scalable.
Figure~\ref{fig:lightgmb} provides a visual overview of the LightGBM architecture.

\subsection{D. Long Short-Term Memory (LSTM)}

LSTM is a type of Recurrent Neural Network (RNN) designed to model long-term dependencies in sequential data. It uses gating mechanisms to control the flow of information. The internal operations of an LSTM cell are governed by the following equations:

\begin{align}
i_t &= \sigma(W_i x_t + U_i h_{t-1} + b_i) \\
\tilde{C}_t &= \tanh(W_C x_t + U_C h_{t-1} + b_C) \\
f_t &= \sigma(W_f x_t + U_f h_{t-1} + b_f) \\
C_t &= f_t \odot C_{t-1} + i_t \odot \tilde{C}_t \\
o_t &= \sigma(W_o x_t + U_o h_{t-1} + b_o) \\
h_t &= o_t \odot \tanh(C_t)
\end{align}

Here, $i_t$, $f_t$, and $o_t$ represent the input, forget, and output gates, respectively. $\tilde{C}_t$ is the candidate cell state, $C_t$ is the actual cell state, and $h_t$ is the hidden state. The operator $\odot$ denotes element-wise multiplication.

\begin{figure}[h]
  \centering
  \includegraphics[width=0.45\textwidth]{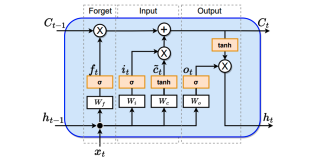}
  \caption{LSTM Cell Diagram}
  \label{fig:lstm}
  \cite{hochreiter1997long}
\end{figure}

LSTM models are typically trained using Backpropagation Through Time (BPTT) and optimized with algorithms such as Adam. Their strength lies in capturing complex temporal relationships, making them ideal for time-series forecasting tasks. Figure~\ref{fig:lstm} illustrates the internal architecture of an LSTM cell.\cite{hochreiter1997long}

\subsection{D. Bidirectional Long Short-Term Memory (LSTM)}
(BiLSTM) is an extension of the traditional long-short-term memory (LSTM) network that can improve model performance on sequence prediction problems. Instead of processing the data only from past to future (forward direction), it also processes from future to past (backward direction), thus capturing context from both directions \cite{schuster1997bidirectional}.

\vspace{1em}
Each BiLSTM consists of two LSTM layers:
\begin{itemize}
    \item A \textbf{forward LSTM} that processes the sequence from $t = 1$ to $t = T$
    \item A \textbf{backward LSTM} that processes the sequence from $t = T$ to $t = 1$
\end{itemize}

At each time step $t$, the hidden states from both directions are concatenated:
\[
\overleftrightarrow{h}_t = [\overrightarrow{h}_t; \overleftarrow{h}_t]
\]

\section*{LSTM Cell Formulas}

The LSTM cell at time step $t$ (for both forward and backward directions) computes its internal states as follows:
\begin{figure}[h]
  \centering
  \includegraphics[width=0.45\textwidth]{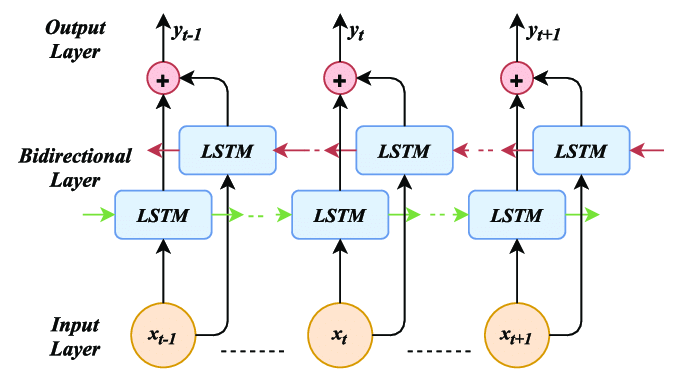}
  \caption{BiLSTM Cell Diagram}
  \label{fig:bilstm}
\end{figure}

This section lays the foundation for comparing the performance, interpretability, and explainability of each model in the context of mid-term electricity load forecasting.

\section{Evaluation Metrics}

All four models of these models above-Linear Regression (LR), XGBoost (XGB), LightGBM, and Long Short-Term Memory (LSTM)—were trained using historical data from January 2016 to December 2023. All models were evaluated on the same four key performance metrics below which help used indicate prediction accuracy. They are calculated as below:

\begin{itemize}

    \item \textbf{Mean Absolute Error (MAE):}
    \[
    \text{MAE} = \frac{1}{n} \sum_{i=1}^{n} \left| y_i - \hat{y}_i \right|
    \]
    MAE measures the average absolute difference between predicted load $\hat{y}_i$ and actual load $y_i$ over $n$ time steps. It treats all errors equally and is particularly useful when outliers are not expected to dominate.
    \newline\item \textbf{Root Mean Squared Error (RMSE):}
    \[
    \text{RMSE} = \sqrt{ \frac{1}{n} \sum_{i=1}^{n} \left( y_i - \hat{y}_i \right)^2 }
    \]
    RMSE penalizes larger errors more than MAE by squaring the residuals before averaging. This metric is more sensitive to large deviations and is often used when the consequences of under- or over-predicting are severe.
    \newline
    \item \textbf{Mean Absolute Percentage Error (MAPE):}
    \[
    \text{MAPE} = \frac{100}{n} \sum_{i=1}^{n} \left| \frac{y_i - \hat{y}_i}{y_i} \right|
    \]
    MAPE expresses prediction error as a percentage of the actual values. It provides scale-independent insight but becomes unstable when actual values $y_i$ approach zero, making it less reliable during low-demand periods.
    \newline
    \item \textbf{Peak-Specific MAPE:}
    \[
    \text{Peak-MAPE} = \frac{100}{k} \sum_{i \in \mathcal{P}} \left| \frac{y_i - \hat{y}_i}{y_i} \right|
    \]
    where $\mathcal{P}$ denotes the set of time indices corresponding to the top 5\% of observed hourly demand, and $k = |\mathcal{P}|$ is the number of peak intervals. Peak-MAPE isolates model accuracy during high-demand periods, which are often the most operationally critical and error-prone. By focusing on extreme values, it evaluates how well a model can anticipate spikes driven by weather or behavioral anomalies.
    \newline
    \item \textbf{SHAP: SHapley Additive exPlanations):}

We employ SHAP (SHapley Additive exPlanations), a framework based on  game theory. SHAP assigns each feature a numerical value based on importance for a particular prediction by computing the average marginal contribution of that feature across all possible feature combinations. This method is grounded through a concept regularly used  in game theory, which ensure a fair distribution of contribution among players (features) in a coalition (the model's prediction).

Given a model \( f \), an input instance \( x \), and a set of all features \( F \), the SHAP value \( \phi_i \) for feature \( i \) is defined as:

\begin{equation}
\phi_i = \sum_{S \subseteq F \setminus \{i\}} \frac{|S|! (|F| - |S| - 1)!}{|F|!} \left[ f(S \cup \{i\}) - f(S) \right]
\end{equation}

Here, \( S \) represents a subset of features not including \( i \), and \( f(S) \) denotes the model's prediction using only the features in subset \( S \). This formulation ensures consistency and local accuracy, making SHAP a robust tool to understand individual predictions and the general behavior of the model. In our context, SHAP helps identify which variables (e.g., lagged load, temperature) most significantly affect the forecast electricity demand.
\end{itemize}

\section{Training and Testing Strategy}

To explore effective strategies for mid-term electricity load forecasting, this study implemented structured training on a variety of machine learning models and engineered feature sets. Our data set included hourly electricity load and weather data for the SCENT region from 2016 through 2024. Data were split chronologically as it was trained in past years and tested on unseen future periods to mimic real-world deployment and avoid data leakage.

We trained several models, including Linear Regression, LightGBM, XGBoost, and LSTM. Each model was first trained with a standard feature baseline comprising calendar features (e.g., \texttt{hour}, \texttt{dayofweek}, \texttt{month}) and weather variables (e.g., \texttt{tavg}, \texttt{tmin}, \texttt{tmax}, \texttt{prcp}). From there, we followed an iterative,  cycle using SHAP to evaluate and improve model inputs.

SHAP was applied after each model training phase to understand which features contributed the most to the predictions and which were considered useless, particularly during high-demand periods. This analysis motivated the engineering of additional features tailored toward capturing peak load behavior—such as \texttt{load\_spike\_vs\_mean}, \texttt{is\_extreme\_heat\_event}, and \texttt{tmin\_roll\_min\_72}. We retrained models after introducing these peak-sensitive features and used SHAP again to verify their impact on predictions to make sure we understood which features were contributing the most. We also continued to test the MAPE as well as other statistical indicators of accuracy before and after to understand whether our newly engineered features were contributing to overall model accuracy or not.

To assess whether models were learning peak behavior effectively, we performed \textbf{targeted testing during known peak load events}. Specifically, we sliced the test set to isolate time intervals of historical winter and summer extremes usually in January and August and re-evaluated SHAP values and prediction accuracy within these subwindows. This allowed us to observe whether added features helped models adapt to unusual patterns of demand driven by extreme temperatures or holidays. We were able to see which features were causing us to underpredict peak values and found ways to refine them by making thresholds to prevent underprediction.

Our iterative refinement strategy included:
\begin{itemize}
    \item Training base models on  calendar and weather features
    \item Engineering lagged, rolling, and interaction terms inspired by SHAP
    \item Conducting ablation tests to remove underperforming features
    \item Evaluating SHAP explanations on full-year and peak-specific intervals
    \item Visualizing prediction curves to diagnose underprediction during peak seasons
    \item Repeating this process to refine both feature sets and model architectures
\end{itemize}

All models were trained on data from 2016--2023 and tested on 2024. Testing performance was always evaluated both \textbf{globally} (on all of 2024 YTD) and \textbf{locally} (on hand-selected peak periods) to ensure that model improvements were meaningful in both average-case and worst-case demand conditions.

 Our approach emphasized adaptive model development guided by both empirical performance and SHAP-based feature explainability. This ensured that feature additions were only contributing directly to our predictions, but we were also able to understand how certain features would cause the model to under predict different peak values.

\section{DATA DESCRIPTION}

This study used hourly electricity load data from ERCOT’s SCENT region through publicly available on ERCOT's website. All data was in the Central Timezone and was across 2016 to 2025 To model the effect of environmental factors on electricity demand, we integrated hourly meteorological data from NOAA. However, since a unified weather dataset does not exist for the SCENT region as a whole, we extracted weather data from four representative cities—Austin, San Antonio, Round Rock, and San Marcos—chosen based on their geographic coverage and relative population within the region.

To approximate region-level weather effects, we applied a population-weighted averaging scheme for each weather variable. Specifically, we assigned a weight to each city proportional to its population size, then computed a weighted sum across cities at each hourly timestamp. This approach ensured that cities with higher electricity demand, implied by greater population density, had a stronger influence on the resulting temperature, precipitation, and snow metrics. This method helped reduce localized anomalies while better reflecting the aggregate climate experienced by SCENT consumers.

The weather data was matched with the load data’s time zone. To ensure timestamp alignment, we resampled all time series to hourly intervals we made sure daylight saving as well as leap years were accounted for in our data.Missing weather values were filled using linear interpolation for small gaps and forward-fill methods for longer, yet short-term gaps. Additionally, we applied additional checks to ensure there were no missing values, or duplicate values.

In addition to the raw temperature and precipitation data, we engineered time-based calendar features such as \texttt{hour}, \texttt{dayofweek}, and \texttt{month}, along with  indicators for weekends and individual weekdays represented by a 0 and 1. We also used holiday effects using the \texttt{USFederalHolidayCalendar} from the pandas t-series module. This allowed us to generate a binary \texttt{is_holiday} feature. as well as interaction terms like \texttt{holiday_x_hour} and \texttt{is_holiday_and_cold} which ended up being particularly relevant when capturing demand outliers during federal holidays and extreme weather conditions like Christmas for example.

Finally,The final dataset consisted of both raw observations and engineered features, all indexed by a continuous hourly time series. This cleaned, consolidated dataset formed the foundation for all modeling and analysis conducted in this study.

\begin{figure}[htbp]
    \centering
    \includegraphics[width=0.9\linewidth]{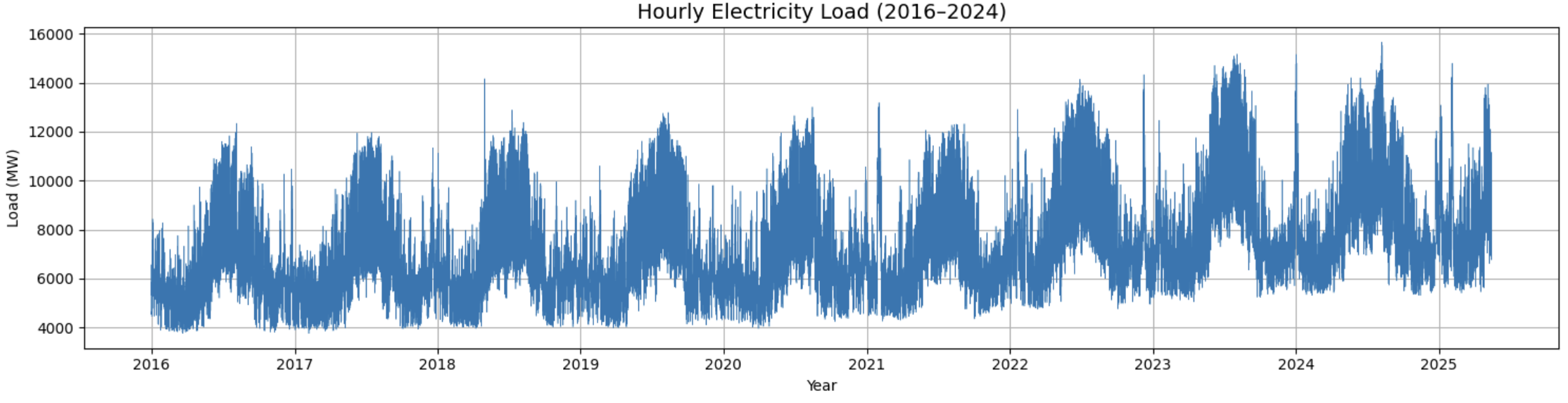} 
    \caption{Load Forecasting Data in ERCOT's SCENT region from 2016 to 2025}
    \label{fig:load_patterns}
\end{figure}

\section{Results and Analysis}
\subsection*{Collinearity Consideration and Feature Distinctiveness}

While some of the input features employed in the model have high statistical correlation, each was specifically designed to capture a distinct behavioral, temporal, or non-linear characteristic of electricity demand. Their presence is justified using SHAP value analysis, which shows that even highly correlated features can provide non-redundant and distinct signals that increase the forecasting accuracy and interpretability of the model.

Lagged Load Features:
Features such as load_lag_24, load_roll_mean_24, and load_roll_max_24 are all derived from recent load history but with varying aims:

load_lag_24 uses the exact load value at the same hour a day earlier. This enables the model to pick up daily seasonality and match hour-to-hour usage.

load_roll_mean_24 provides a smoothed view of the past 24 hours, which helps with noise reduction as well as understanding underlying demand.

load_roll_max_24 provides the peak load that occurred in the previous day, which is especially relevant during peak periods such as hot afternoons or cold mornings.

Although these features are related, each allows the model to detect and respond to different load patterns — a repeating cycle, a steady trend, or a short-term spike.

Temperature Features:
Similarly, features like tmax, tavg_lag_24, and tmax_lag_24 all derive from weather data but measure distinct temperature influences:
tmax gives the current day's outdoor maximum temperature and is  associated with air conditioning. tavg_lag_24 is the previous day's average temperature, which accounts for the lagged effect of thermal mass of buildings and occupant behavior.

Model Behavior and Redundancy Handling:
Tree-based models like XGBoost handle multicollinearity by design. They do not assign coefficients to all of the features like linear regression models. Instead, they greedily choose one feature at a time to split on — the one that most enhances the model's objective function at the present step. Even if two features are highly correlated, the model can learn to prefer one in some situations and the other in others. 

\clearpage
\begin{table}[p]
\centering
\small
\centering
\small
\caption{Top SHAP Features: Detailed Calculation and Forecasting Importance}

\begin{adjustbox}{width=\textwidth}
\begin{tabular}{|p{4cm}|p{6cm}|p{6cm}|}
\hline
\textbf{Feature} & \textbf{How It's Calculated} & \textbf{Why It's Important} \\
\hline
\texttt{load\_roll\_mean\_24} & 24-hour rolling mean of historical load, shifted by 1 hour to prevent leakage. & Smooths hourly noise and provides a short-term demand trend reference, aiding in near-future predictions. \\
\hline
\texttt{load\_spike\_vs\_mean} & Normalized deviation: (current load – 24h rolling mean) / (rolling mean + 1). & Highlights abnormal load increases relative to recent history, useful for detecting extreme peaks. \\
\hline
\texttt{load\_lag\_24} & The load recorded exactly 24 hours before the current timestamp. & Captures daily seasonality, especially relevant in residential and commercial electricity usage patterns. \\
\hline
\texttt{load\_roll\_max\_24} & The maximum load value over the past 24 hours (lagged by 1 hour). & Represents recent peak demand conditions; important during high-usage periods like summer afternoons. \\
\hline
\texttt{tmax} & Daily maximum temperature reported at the hour level. & Closely correlated with air conditioning use; vital for forecasting load on hot days. \\
\hline
\texttt{tavg\_lag\_24} & The average temperature from 24 hours ago. & Helps model delayed or cumulative thermal effects on usage, especially in residential buildings. \\
\hline
\texttt{CDD} & Cooling Degree Days: max(0, tavg – baseline). & Measures accumulated heat exposure; tracks increasing A/C load during hot seasons. \\
\hline
\texttt{tmax\_lag\_24} & Daily max temperature from 24 hours prior. & Captures lagged heat influence—useful when building thermal inertia affects next-day demand. \\
\hline
\texttt{load\_roll\_std\_24} & Standard deviation of load over the last 24 hours (shifted). & Quantifies recent variability or demand volatility, flagging potential instability. \\
\hline
\texttt{hour} & Extracted hour of day (0–23) from timestamp. & Encodes daily cyclical behavior—helps the model recognize morning vs. afternoon vs. night load differences. \\
\hline
\texttt{load\_roll\_mean\_336} & 336-hour (14-day) rolling average of load (lagged). & Provides medium-term load trend context and helps filter out weekly anomalies. \\
\hline
\texttt{dayofweek} & Integer from 0 (Monday) to 6 (Sunday) extracted from timestamp. & Captures behavioral shifts between weekdays and weekends, especially in industrial/residential zones. \\
\hline
\texttt{tmin} & Minimum daily temperature recorded. & Useful for modeling heating or morning load during cold weather periods. \\
\hline
\texttt{load\_lag\_48} & Load value from exactly 48 hours ago. & Reinforces two-day demand cycles (e.g., Tuesday = Sunday pattern), useful for calendar-heavy effects. \\
\hline
\texttt{CDD\_x\_hour} & Interaction term: Cooling Degree Days multiplied by the hour. & Captures non-linear peaks in cooling demand, typically in mid-to-late afternoons. \\
\hline
\texttt{load\_lag\_168} & Load value from 168 hours (7 days) ago. & Encodes weekly seasonality; helps recognize long-term behavioral repetition (e.g., every Friday). \\
\hline
\texttt{load\_roll\_std\_168} & Standard deviation over past 168 hours (shifted by 1 hour). & Captures week-long load fluctuation patterns; critical for capturing unpredictable weekly changes. \\
\hline
\texttt{tmin\_lag\_24} & Minimum temperature recorded 24 hours prior. & Models delayed heating effects, especially for early morning peaks following cold nights. \\
\hline
\texttt{month} & Extracted month number (1–12) from timestamp. & Encodes broad seasonal effects like summer/winter that drive major changes in load. \\
\hline
\texttt{temp\_spike\_vs\_mean} & (tmax – tavg) / (tavg + 1): measures how extreme max temp is relative to average. & Detects unusual heat spikes, which often lead to extreme load changes, especially during heatwaves. \\
\hline
\end{tabular}
\end{adjustbox}
\label{tab:shap_features}
\end{table}
\clearpage
\subsection{Linear Regression Results}
We began by training a baseline linear regression model using standard calendar, weather, and lagged load features. This model achieved a mean absolute percentage error (MAPE) of 5.37\% on 2024 load data. Although the baseline captured general patterns, it struggled with peak events and context-sensitive fluctuations.

To understand and refine the model, we used SHAP (SHapley Additive exPlanations) to assess the contribution of each feature. SHAP analysis showed that lag-based variables such as \texttt{load_lag_24}, \texttt{load_roll_mean_24}, and \texttt{load_lag_168} had the highest impact on predictions, while static calendar indicators contributed far less.

Motivated by these insights, we enhanced the feature set by introducing dynamic variables that better captured temporal and weather-related anomalies. These included \texttt{CDD_lag_24}, \texttt{HDD_lag_24}, \texttt{temp_spike_vs_mean}, \texttt{is_extreme_cold_event}, \texttt{is_extreme_heat_event}, as well as cyclical encodings such as \texttt{hour_sin} and \texttt{dayofweek_cos}. We also engineered interaction terms like \texttt{lag_24_x_hour} and \texttt{CDD_x_hour}.

This refined model achieved a significantly improved MAPE of 4. 74\%, a relative reduction of more than 11\%. The enhanced performance demonstrates how SHAP can serve as a guide to enhanced results.

\subsection{SHAP-Guided Feature Engineering for XGBoost}

The baseline XGBoost model, trained on traditional features such as weather conditions (\texttt{tavg}, \texttt{tmax}, \texttt{tmin}), calendar variables (\texttt{month}, \texttt{dayofweek}), and lagged load values (\texttt{load\_lag\_24}, \texttt{load\_lag\_168}), achieved a moderate MAPE of 3.21\%. However, SHAP value analysis revealed that the model consistently underpredicted during peak demand events—particularly early morning winter peaks and late afternoon summer peaks.

Calendar features like \texttt{month} and \texttt{dayofweek} dominated the global SHAP importance plots, yet their static nature failed to account for dynamic weather-driven spikes. For instance, during heatwaves, the model underestimated load despite high \texttt{tmax} values because the baseline features lacked sensitivity to temperature volatility and interaction with usage patterns. Similarly, sharp morning ramps in winter went undetected due to the absence of early-hour volatility indicators. We found long term trends specifically like the month variable which keeps causes the forecast to under predict the peak points because historically January had a lower load demand than other months. This caused the forecast to under predict spikes as long-term variables would keep predictions low regardless of spikes.

To address this, we engineered a new set of features guided by SHAP diagnostics:
\begin{itemize}
    \item \textbf{Spike detection features:} \texttt{load\_spike\_vs\_mean} is calculated as the percentage deviation of the current load from its 24-hour rolling mean:
    \[
        \texttt{load\_spike\_vs\_mean} = \frac{\texttt{load} - \texttt{load\_roll\_mean\_24}}{\texttt{load\_roll\_mean\_24} + 1}
    \]
    This feature captures sudden increases or drops in demand, which helps the model better detect unexpected surges often seen during peak periods.
    \newline
    \item \textbf{Temperature volatility:} \texttt{temp\_spike\_vs\_mean} measures the relative difference between the day's maximum temperature and the average temperature:
    \[
        \texttt{temp\_spike\_vs\_mean} = \frac{\texttt{tmax} - \texttt{tavg}}{\texttt{tavg} + 1}
    \]
    In addition, \texttt{tmax\_roll\_max\_72} computes the rolling 3-day (72-hour) maximum of \texttt{tmax}. These features detect heatwaves or rapid temperature increases that typically drive peak loads due to air conditioning use.
    \newline
    \item \textbf{Interaction terms:} \texttt{CDD\_x\_hour} multiplies the cooling degree days (CDD) with the hour of day:
    \[
        \texttt{CDD\_x\_hour} = \texttt{CDD} \times \texttt{hour}
    \]
    Likewise, \texttt{lag\_24\_x\_hour} multiplies the 24-hour lagged load with the hour:
    \[
        \texttt{lag\_24\_x\_hour} = \texttt{load\_lag\_24} \times \texttt{hour}
    \]
    These interaction terms allow the model to learn how the influence of weather and past demand varies by time of day, which is especially relevant during recurring peak hours (e.g., 5–7 PM).
    \newline
    \item \textbf{Extreme event flags:} \texttt{is\_extreme\_heat\_event} is a binary flag set to 1 when the \texttt{tmax} value exceeds the 95th percentile, marking rare but high-impact heat events. \texttt{is\_monday} flags the start of the workweek when sudden shifts in demand often occur. Both features provide explicit signals for outlier behavior in the load pattern that the model would otherwise struggle to infer.
\end{itemize}

After integrating these features, the improved XGBoost model achieved a significantly lower MAPE of 0.79\%. SHAP summary plots of the refined model revealed a shift in importance: static calendar features were replaced by dynamic load- and weather-sensitive predictors. Notably, \texttt{load\_spike\_vs\_mean} and \texttt{CDD\_x\_hour} ranked among the top contributors, indicating that the model now prioritized real-time signals over general trends. This transformation demonstrates the power of explainability not only to interpret model behavior but also to guide tangible performance improvements—particularly for better peak load forecasting.

\subsection*{SHAP-Guided Feature Engineering for LightGBM}

The baseline LightGBM model relied on traditional calendar features (\texttt{hour}, \texttt{dayofweek}, \texttt{month}) and basic weather/load variables (\texttt{tmax}, \texttt{tavg}, \texttt{load\_lag\_24}, \texttt{load\_lag\_168}). While it achieved moderate performance with a MAPE of 5.26\%, it consistently underpredicted electricity load during high-demand periods. As seen in the baseline SHAP summary plot, the top predictors lacked direct representations of peak sensitivity, temperature volatility, and temporal interaction effects.

To address these shortcomings, we applied SHAP (SHapley Additive exPlanations) analysis to pinpoint where the model was under performing and identify feature gaps. This led to the design of new engineered features specifically targeting peak conditions:

\begin{itemize}
    \item \textbf{Spike Detection:} 
    \[
        \texttt{load\_spike\_vs\_mean} = \frac{\texttt{load} - \texttt{load\_roll\_mean\_168}}{\texttt{load\_roll\_std\_168} + \epsilon}
    \]
    This feature captured abrupt deviations from typical load behavior and became a key driver of peak prediction accuracy.
    \newline
    \item \textbf{Temperature Volatility:}
    \[
        \texttt{temp\_spike\_vs\_mean} = \texttt{tmax} - \texttt{tavg}
    \]
    Sudden spikes in temperature were associated with increased HVAC load and were critical in anticipating afternoon peaks.
    \newline
    \item \textbf{Temporal Interactions:}
    \[
        \texttt{lag\_24\_x\_hour} = \texttt{load\_lag\_24}\times\texttt{hour}
    \]
    These features allowed the model to account for how past load and cooling demand varied by time of day, capturing diurnal patterns missed by baseline predictors.
    \newline
    \item \textbf{Extreme Event Flag:}
    \[
        \texttt{is\_extreme\_heat\_event} = \mathbb{1}[\texttt{tmax} > Q_{95}]
    \]
    A binary indicator identifying heatwaves helped differentiate high-demand anomalies from typical warm days.
\end{itemize}

The improved SHAP summary plot demonstrated a significant shift in feature importance: engineered features such as \texttt{load\_spike\_vs\_mean}, \texttt{load\_roll\_mean\_168}, and \texttt{CDD\_x\_hour} became dominant, while static calendar features lost relevance. This shift reflected a model better attuned to short-term dynamics and non-linear interactions.

\textbf{Performance gains} were substantial:
\begin{itemize}
    \item MAPE improved from 5.26\% to 0.98\%.
    \item Peak MAPE (top 5\% load hours) dropped to 0.79\%.
    \item MAE fell from 447.48 to 85.03.
\end{itemize}

\subsection*{LSTM and Bidirectional LSTM Modeling}

To benchmark against gradient boosting models, we implemented a Long Short-Term Memory (LSTM) network to capture temporal dependencies in the load data. SHAP was not used in this context, as Deep SHAP and gradient-based explainer produced unstable results due to the model's recurrent architecture.

Instead, we drew upon prior feature engineering insights from tree-based models—such as \texttt{load\_lag\_24}, \texttt{load\_roll\_mean\_24}, and \texttt{temp\_spike\_vs\_mean}—and incorporated them as input sequences to the LSTM.

We then extended the architecture to a Bidirectional LSTM, allowing the model to learn patterns from both past and future contexts within the input window. This architectural refinement improved performance, reducing the MAPE from 1.21\% to 1.14\% on the 2024 test set. Despite the modest gain, this demonstrated the value of temporal context and feature-informed design in deep learning-based load forecasting.

\begin{table}[h!]
\centering
\caption{Model Performance Comparison (2024)}
\label{tab:model_performance}
\begin{tabular}{lccc}
\toprule
\textbf{Model} & \textbf{RMSE} & \textbf{MAE} & \textbf{MAPE } \\
\midrule
Linear Regression (Baseline)       & 658.59 & 459.01 & 5.37 \\
Linear Regression (Improved)       & 559.46 & 396.01 & 4.74 \\
XGBoost (Baseline)                 & 421.94 & 273.51 & 3.15 \\
XGBoost (Improved with SHAP)       & \textbf{106.17} & \textbf{70.43} & \textbf{0.79} \\
LightGBM (Baseline)                & 447.48 & 579.58 & 5.26 \\
LightGBM (Improved with SHAP)      & 136.38 & 85.03  & 0.91 \\
LSTM                               & 193.66 & 101.32 & 1.21 \\
Bidirectional LSTM                 & 191.48 & 95.66  & 1.14 \\
\bottomrule
\end{tabular}
\end{table}

To add, we noticed that the features we used seemed to have a high collinearity after running tests but after removing the features that did have a high collinearity it seemed to reduce perfomance immensely. After testing we noticed that while this may affect feature importance that SHAP provides collinearity had not negative affect on the performance itself for XGBoost. For future works it may be helpful to instead of dropping features with high collinearity to combine them therefore performance will hopefully not reduce while still being able to get transparent feature values
\section{Conclusions}
Through a systematic comparison of machine learning models for electricity load forecasting, our results highlight the power of explainability-driven feature engineering. SHAP analysis on the XGBoost and LightGBM models revealed that traditional calendar and weather features (e.g., \texttt{month}, \texttt{dayofweek}, \texttt{is_holiday}, \texttt{tavg}) alone were insufficient in capturing high-load events, leading to consistent under prediction during extreme peaks.

To address this, we engineered several targeted features:
\begin{itemize}
\item \texttt{load_spike_vs_mean} improved sensitivity to short-term surges.
\item \texttt{tmax_roll_max_72} and \texttt{temp_spike_vs_mean} captured sustained heatwave conditions.
\item \texttt{lag_24_x_hour} and \texttt{CDD_x_hour} enabled temporal interactions between load and temperature.
\item \texttt{is_extreme_heat_event} flagged rare demand spikes due to unusual temperature thresholds.
\end{itemize}

These features dramatically improved peak forecasting accuracy. XGBoost, in particular, achieved a MAPE of just \textbf{0.79\%} on 2024 data, with a similar performance from LightGBM after integrating SHAP-driven insights. While deep learning approaches like LSTM and Bidirectional LSTM showed strong temporal modeling capabilities (MAPE down to \textbf{1.14\%}), they lacked direct interpretability and did not benefit from SHAP due to limitations with Deep SHAP explainability.

In summary, XGBoost with SHAP-informed feature engineering provided the best overall accuracy and explainability, making it the most effective and explainable model for both average and extreme load forecasting scenarios. It also showed that SHAP informed feature engineering is a valuable tool not only for explainability but also for predicting peaks and improving performance.
\clearpage
\section{Models}
\begin{figure}[htbp]
    \centering
    \includegraphics[width=1.05\textwidth]{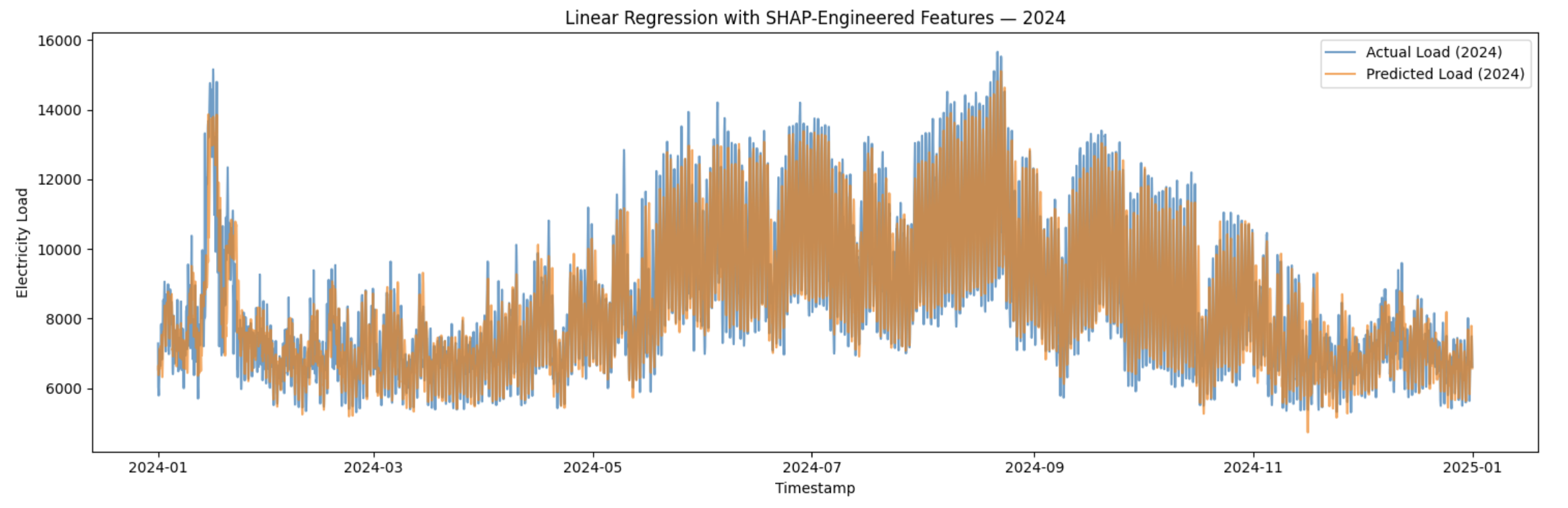} 
    \caption{Improved Linear Regression Forecast}
    \label{fig:load_patterns2}
\end{figure}

\begin{figure}[htbp]
    \centering
    \includegraphics[width=1.05\textwidth]{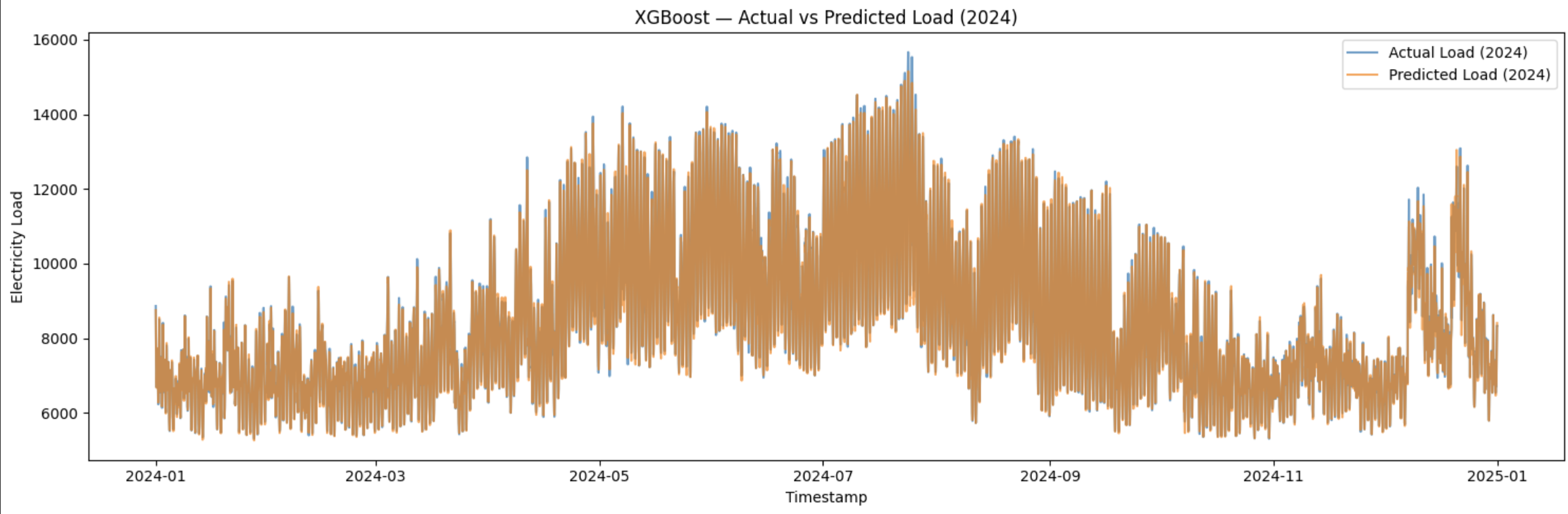} 
    \caption{Improved XGBoost Regression Forecast}
    \label{fig:load_patterns2}
\end{figure}

\begin{figure}[htbp]
    \centering
    \includegraphics[width=1.05\textwidth]{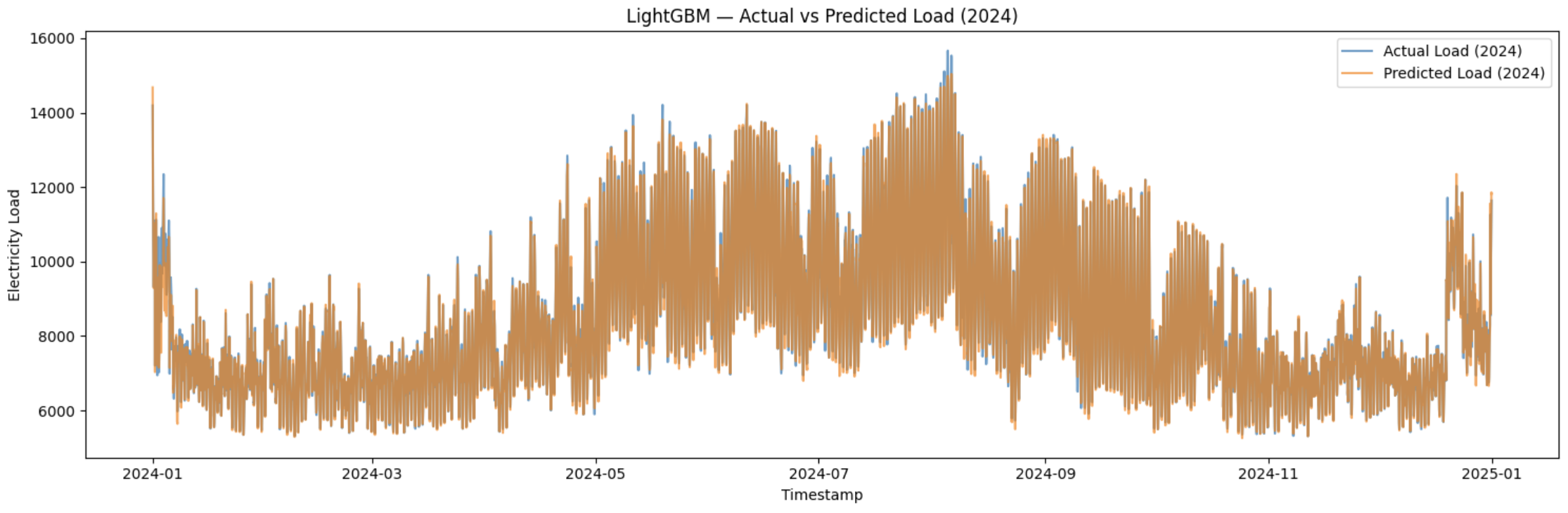} 
    \caption{Improved LightGBM Regression Forecast}
    \label{fig:load_patterns}
\end{figure}

\clearpage

\clearpage
\begin{figure}[htbp]
    \centering
    \includegraphics[width=1\linewidth]{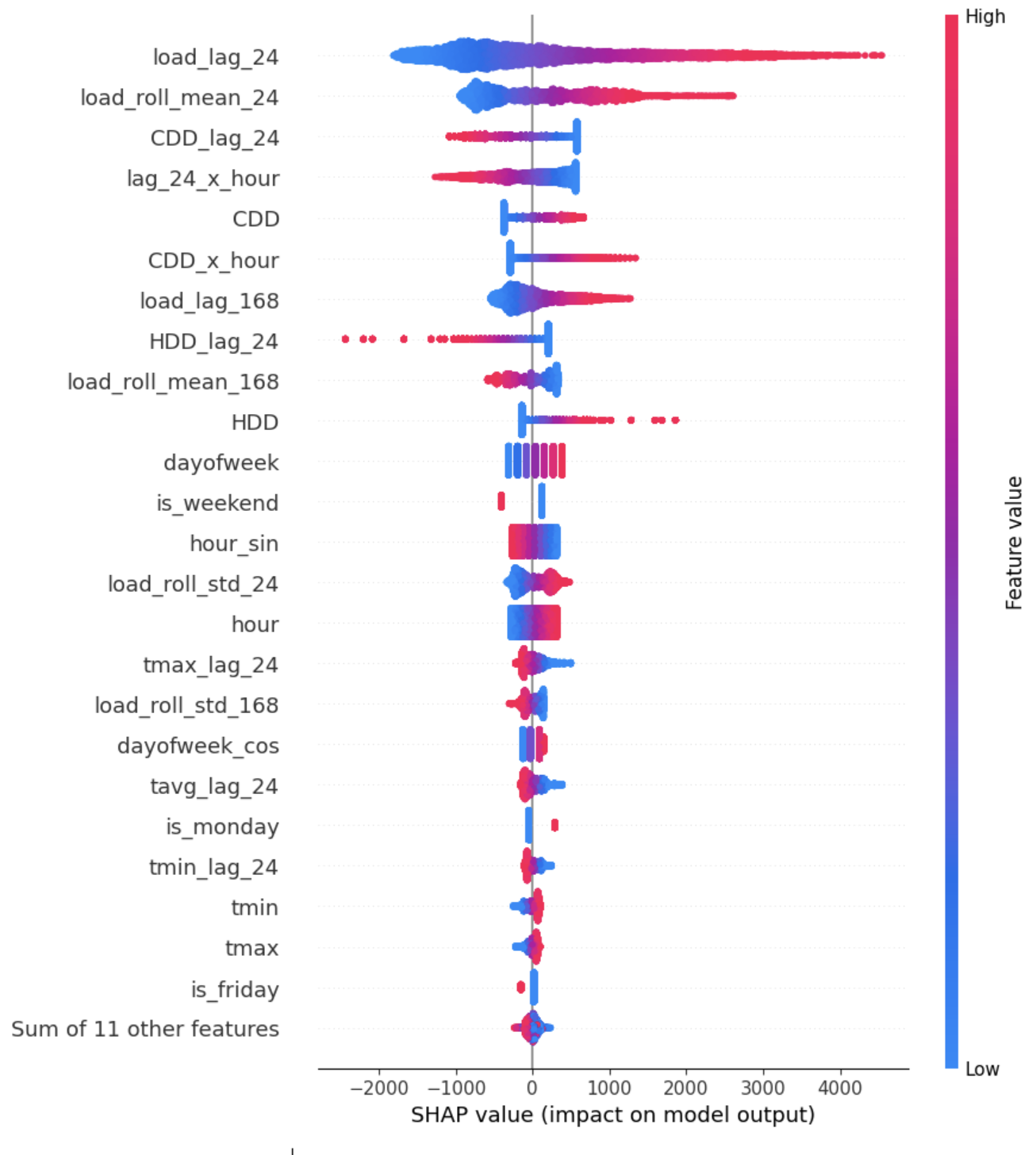} 
    \caption{SHAP analysis of Linear Regression Model}
    \label{fig:load_patterns3}
\end{figure}

\begin{figure}[htbp]
    \centering
    \includegraphics[width=1\linewidth]{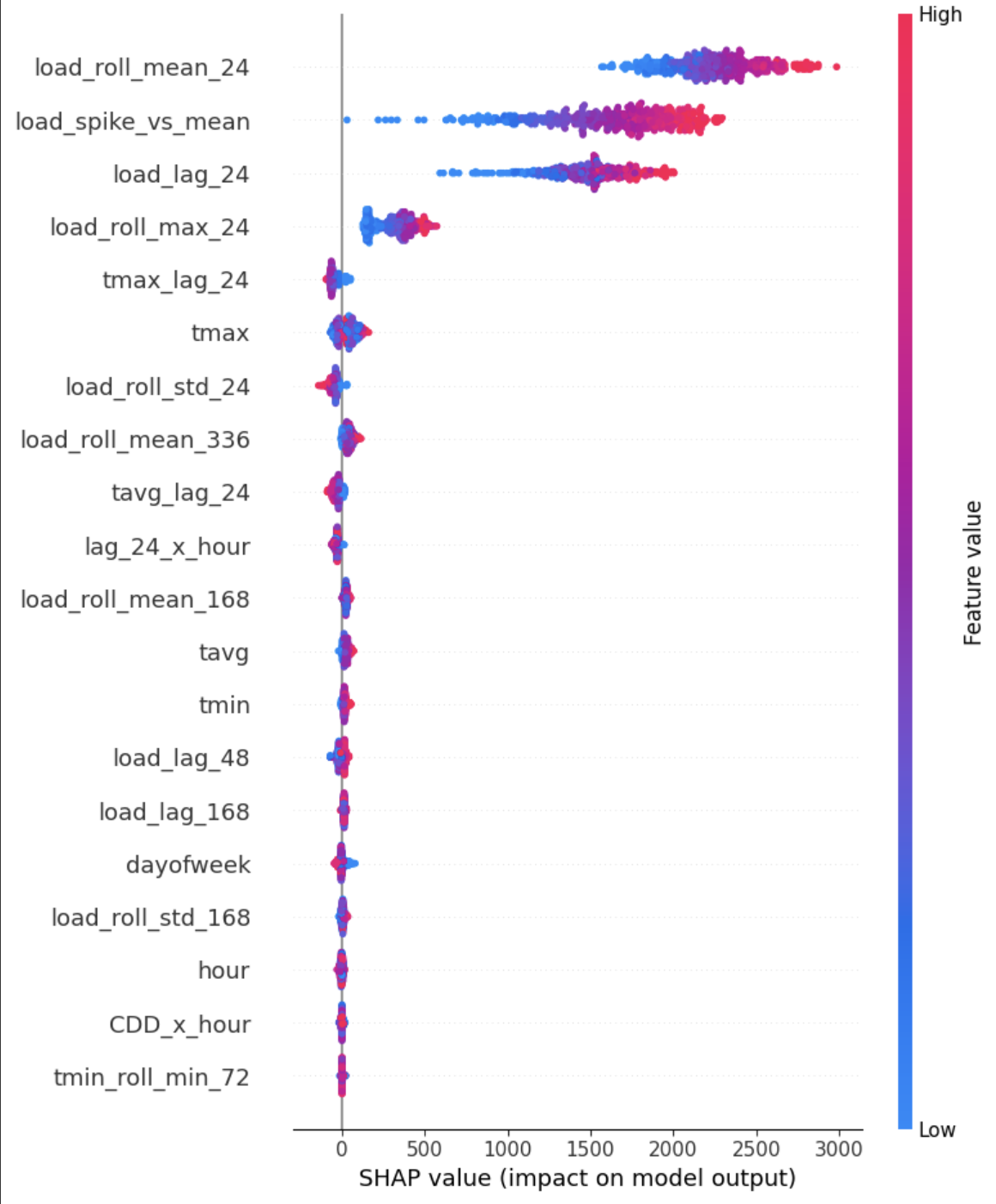} 
    \caption{SHAP analysis of improved XGBoost Model}
    \label{fig:load_patterns}
\end{figure}

\begin{figure}[htbp]
    \centering
    \includegraphics[width=1\linewidth]{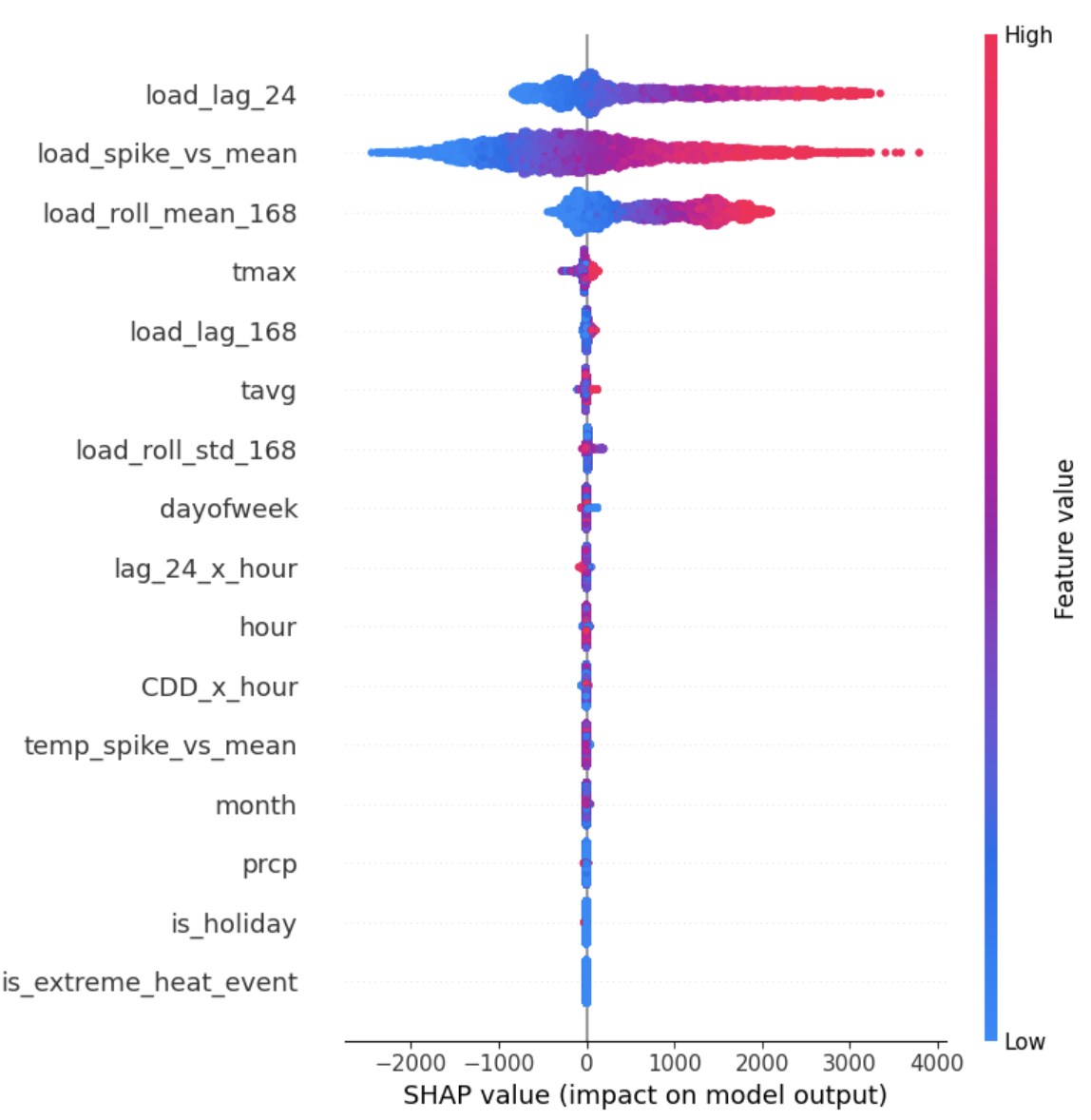} 
    \caption{SHAP analysis of improved LightGBM Model}
    \label{fig:load_patterns}
\end{figure}

\bibliography{reference}

\end{document}